\documentclass[sigconf]{acmart}

\usepackage{booktabs} 
\usepackage{mathtools}
\usepackage{algorithm2e}
\usepackage{graphicx}
\usepackage{subfigure}

\setcopyright{rightsretained}

\acmDOI{10.1145/3321707.3321828}

\acmISBN{978-1-4503-6111-8/19/07}

\acmConference[GECCO '19]{the Genetic and Evolutionary Computation Conference 2019}{July 13--17, 2019}{Prague, Czech Republic}
\acmYear{2019}
\copyrightyear{2019}

\acmPrice{15.00}

\begin{document}
\title{Lexicase Selection in Learning Classifier Systems}

\author{Sneha Aenugu}

\affiliation{%
  \institution{University of Massachusets Amherst}
  \streetaddress{}
  \city{Amherst} 
  \state{Massachusetts} 
  \postcode{}
}
\email{saenugu@umass.edu}

\author{Lee Spector}

\affiliation{%
\institution{Hampshire College}
  \institution{University of Massachusetts Amherst}
  \streetaddress{}
  \city{Amherst} 
  \state{Massachusets} 
  \postcode{}
}
\email{lspector@hampshire.edu}

\renewcommand{\shortauthors}{S. Aenugu et al.}

\begin{abstract}
  The lexicase parent selection method selects parents by considering performance on individual data points in random order instead of using a fitness function based on an aggregated data accuracy. While the method has demonstrated promise in genetic programming and more recently in genetic algorithms, its applications in other forms of evolutionary machine learning have not been explored. In this paper, we investigate the use of lexicase parent selection in Learning Classifier Systems (LCS) and study its effect on classification problems in a supervised setting. We further introduce a new variant of lexicase selection, called batch-lexicase selection, which allows for the tuning of selection pressure. We compare the two lexicase selection methods with  tournament and fitness proportionate selection methods on binary classification problems. We show that batch-lexicase selection results in the creation of more generic rules which is favorable for generalization on future data. We further show that batch-lexicase selection results in better generalization in situations of partial or missing data.
\end{abstract}

%
%



\maketitle

\section{Introduction}
Genetic Algorithms generally define a fitness function to apportion credit to the candidates in a population, and parent selection methods use this fitness definition to choose the parents that will be used to produce offspring for the next generation. Such fitness functions generally compute the aggregate performance of each candidate across a data set. Particularly in cases in which the training data is heavily biased towards one type of data, the candidates chosen as parents might not be optimal for generalization on future data. This problem becomes relevant in the setting of online-learning or when there is data missing. For increasing future generalization, it is crucial to maintain the diversity and generality of the population in the parent selection step~\cite{Uniq}, ~\cite{Div}. Methods like fitness sharing~\cite{Fshare} and co-solvability~\cite{Cosov} have been developed to mitigate this problem. 

Lexicase selection~\cite{Lex}, by avoiding the use of an aggregating fitness function altogether, can be used to bolster the diversity and generality in  populations~\cite{LexDiv},~\cite{Lex4}. Lexicase selection has been shown to outperform other selection methods in several applications of genetic programming, and in traditional genetic algorithms when applied to Boolean constraint satisfaction problems~\cite{Lex3}. Several variants of lexicase selection have also been proposed for applications in genetic programming~\cite{Lex5},~\cite{Lex6}. A variant of lexicase selection, epsilon lexicase selection ~\cite{Lex2} shows performance improvement in symbolic regression problems with floating point errors. Still, lexicase selection's applications in other forms of machine learning and evolutionary computation remain largely unexplored. This paper aims to study the characteristics of lexicase selection in the context of learning classifier systems.

Learning Classifier Systems (LCS)~\cite{Ryan} are rule-based learning systems that incorporate genetic algorithms to discover rules that characterize a given data set. The main components in LCS include generation of the rule set, fitness assignment, and rule discovery. Traditionally, fitness proportionate selection is used in the parent selection step of LCS. But various studies have shown that tournament selection is more robust to fitness scaling and hyperparameter tuning~\cite{Tourn1}, ~\cite{Tourn2},~\cite{Tourn3}. Tournament selection is also shown to be more robust to noise~\cite{Robust}.  

In this paper, we explore lexicase as an alternate parent selection method in LCS. We further introduce a new variant of lexicase selection called batch-lexicase selection that we have deemed more appropriate for LCS applications. We show that the lexicase selection variants produce more generic rule representations than other methods, which also makes them more robust to missing or biased data. Here we make a comparison between the four selection methods: lexicase selection, batch-lexicase selection, tournament selection and fitness proportionate selection. We compare these methods on four binary classification data sets: multiplexer, parity, LED, and car evaluation. We first report results on the complete data sets using all of the parent selection methods. Following this, we conduct additional tests in which incomplete data is provided to the LCS, and we compare the generalization capacities of various methods. We further discuss the sensitivity to parameter tuning of the newly-introduced batch-lexicase selection algorithm.

Section 2 gives a brief overview of the LCS. Section 3 describes the parent selection methods analyzed in this study and introduces the batch-lexicase selection algorithm. Section 4 explains our experiments describing the different problems used in this study and elucidates the results from these experiments. We discuss our findings and conclude in section 5.

\section{Learning Classifier Systems}
A Learning Classifier System (LCS) evolves a population of rules that best describe the data. It makes use of  a genetic algorithm to improve the rule set every iteration. LCSs come in two flavors - Michigan-style systems which evolve a single solution and Pittsburg-style systems which evolve a set of solutions. Michigan-style systems are widely used in the online setting and are the focus of this study. We start from an empty rule set and gradually generate the rules while data samples are passed to the system one at a time. In this study, we employed  the UCS (sUpervised Classifier System) architecture~\cite{UCS},~\cite{UCS2} which is based on the widely studied XCS architecture ~\cite{XCS}  on a diverse set of classification problems. Both UCS and XCS are accuracy-based architectures where fitness is a measure of accuracy. XCS bases its fitness on accuracy in a reinforcement learning setting while UCS defines its fitness function based on accuracy in a supervised setting. The basic components of the UCS architecture are briefly described below.

\subsection{Rule Generation} In this architecture, we start with an empty rule set and gradually keep adding rules as we see the data one at a time. The rules in the UCS have the format, if <condition>, then <class>. The <condition> is generated by randomly masking a few features of the data sample by \lq\#\rq which are interpreted as  \emph{don't care} features. This process is known as covering. This format captures the correlations between different features of the data.

\subsection{Learning/ Updates} In the context of LCS, the learning process is to update the fitness value of the current rule population on seeing a new data point. A data sample is provided to the system during the learning process and a match set (M) is generated from the existing rule population which contain the rules with condition that matches the data. From this match set, a correct set (C) is generated which has the set of rules that predict the class of the data correctly. For each classifier in the match set and the correct set, the number of matches (also known as experience) and number of correct predictions are updated. The accuracy of the classifier is then updated as follows
    \begin{equation}
      \textnormal{Accuracy} = \frac{\textnormal{Number of correct classifications}}{\textnormal{Number of Matches}}
    \end{equation}
    The fitness function is defined as
    \begin{equation}
      \textnormal{Fitness} = \textnormal{Accuracy}^{\nu}
    \end{equation}
    where $\nu$ is the power parameter used to tune the importance of high accuracy when calculating fitness. In the case where the match set is a null, covering is initiated to generate a new rule explaining the given data point. 
    \subsection{Rule Discovery} A Genetic algorithm (GA) is used to discover new rules from the existing rule population. In the current architecture, parent rules are selected from the correct set of the current data sample, whenever the GA component is triggered. A parent selection technique is used to select two parents from among the correct set (traditionally based on the fitness function) and crossover and mutation operations are performed on the selected parents to generate offspring rules. Fitness proportionate selection was used in the earlier LCS but later research favored tournament selection as a parent selection technique to discover the rules.
    \subsection{Rule Compaction}
    Rule compaction enforces the compactness of the rule representation and ensures the generalization capacity of the rules. A subsumption operation is performed at intervals to identify any rules which are a more specific version of a general rule. If that is the case, the specific version of the rule is deleted and the numerosity of the general rule is incremented. Numerosity of a classifier can be interpreted as the number of repetitions of a rule that is maintained in the rule set. A maximum limit is enforced on the number of rules to describe a given dataset and the least fit rules are deleted from the population whenever the size exceeds the specified limit. 
  \subsection{Prediction}
  During the test mode, the class of a given data sample is to be predicted. A match set is generated from the existing rule set and each rule in the match set votes on a class with its vote proportional to its fitness and numerosity. The votes for all the classes are aggregated and the winning class is assigned to the given data sample.

  There are three places in the architecture where the fitness function is used to perform selection of some sort. In the rule discovery module, the fitness function is used to identify the best candidates for selecting the next generation of rules. In the rule deletion module, once the population size exceeds a given threshold, the rules with the least fitness are selected for deletion. Lastly, in the prediction module, the fitness function is used to assign weight for each rule in predicting the class category of an unseen data sample.  In this study, fitness proportionate selection is used as the selection method for the deletion step. The emphasis is placed on studying the influence of different parent selection techniques in the rule discovery component of the LCS and its effect on the overall classification performance.

\section{Parent Selection in LCS}
The parent selection module chooses a set of individuals among a population from which new rules are generated. For a fast convergence to an optimal rule set, it is preferable for the selected individuals to have good generalization ability towards unseen data. Selection pressure denotes the degree to which better individuals are preferred. Pure elitism (always picking the best individuals in terms of a fitness function) is a technique with high selection pressure. Stochastic selection (randomly picking two individuals) on the other hand has very low selection pressure. There are different parent selection techniques with selection pressure in the spectrum with extremes as stochastic selection and pure elitism. Some of the techniques give the flexibility to adapt the selection pressure to fit a given problem. This section briefly describes a few of the commonly employed selection techniques (tournament selection and roulette wheel selection) and other techniques (lexicase and batch-lexicase) which have not been previously explored in the context of LCS and other machine learning applications. Batch-lexicase selection, a variation of lexicase selection which can tune its selection pressure is a contribution of this paper.

\subsection{Fitness Proportionate }
Fitness proportionate selection or the roulette wheel selection is a technique where parents are chosen among the population with a probability proportional to the fitness function. This selection method depends heavily on the definition of the fitness function and was used in the earlier versions of LCS. It was later shown that the performance of fitness proportionate selection is dependent on fitness scaling and the distribution of the fitness function~\cite{Robust}. Some studies, however, contend this claim and show that roulette wheel selection performs competitively with tournament selection if proper fitness separation is employed~\cite{Roul}. The selection pressure in roulette wheel selection is not adjustable.

\subsection{Tournament}

Tournament selection selects the best individual in a randomly picked subset of individuals from a population. The size of the subset picked (tournament size) indicates the selection pressure in this selection. A tournament size of 1 is equivalent to random selection and a tournament size equal to the size of the population is equivalent to the elitist selection. In contrast to roulette wheel selection, tournament selection is independent of fitness scaling as it always chooses the rule with the highest fitness. This was shown to make tournament selection more robust than roulette wheel to hyperparameter settings~\cite{Robust} in LCS. Tournament is widely used as a default selection method in the LCS literature.

\subsection{Lexicase}
Lexicase selection does not base its selection criteria on an error-aggregating fitness function. Instead it gradually eliminates its candidates as it proceeds to look at how the population fares at each data point in a shuffled dataset. In the context of LCS, the rules which match a given randomly picked data point (without replacement) and correctly predict its class are allowed to survive while the rest are eliminated. The process is repeated till there is only one rule left or the set of data samples are exhausted. In the latter case a rule is picked at random from the remaining rule set. In the online setting, data samples which are accumulated till the current iteration are considered in the selection process. The full algorithm is described in Algorithm 1. As in the case of roulette wheel selection, lexicase selection can't tune its selection pressure.

The motivation behind lexicase selection is to allow the survival of those rules which fare low in average but which could fit the data points where a majority of high-achieving rules fail. This could be of potential interest in improving the generalization ability of a model in machine learning and data mining applications. 

\begin{algorithm}
  \KwData{cases := list of training cases in random order, candidates := population of classifiers}
 \KwResult{Return an individual to be used as a parent }
 \While{True}{
  data-sample := first sample from cases\;
  candidates := candidates with the data sample in the correct set\;
  \If{only one candidate exists in candidates}{
  	parent = candidate\;
	return parent;
   }
   delete data-sample from cases; \\
   \If{cases is empty}{
	parent = a randomly selected candidate from candidates\;
	return parent;
    }
   
 }
 \caption{Lexicase Selection}
\end{algorithm}

\subsection{Batch-Lexicase}

The lexicase selection filters test cases gradually by stepping through a randomly ordered set of data points. In situations where each rule represents only a sparse subset of data points, most of the rules are eliminated after navigating only a few data samples. This results in a high variance in the parent selection by being heavily dependent on the ordering of the data points. To overcome this shortcoming, a new variant of the lexicase selection, batch-lexicase selection is introduced. The main idea is to increase the size of the rule population surviving each iteration. In batch-lexicase selection, batches of data are used to filter the rules at each iteration instead of a single data point. The rules which are elite on batches of data are allowed to survive. When the batch size is equal to the entire dataset it is equivalent to the elitist/deterministic selection. The full algorithm for the batch-lexicase selection is described in Algorithm 2. 

Batch-lexicase selection can tune its selection pressure using two hyperparameters. One of them as discussed above is the batch size. The second hyperparameter is the threshold for determining which rules survive a particular dataset. Here, we use the fitness function to indicate the fitness levels of each of the rules when applied to a given batch. The fitness function used here is identical to the definition of accuracy defined in equation (2). A predefined threshold of fitness is used to eliminate the rules in a given batch. If the fitness threshold is low, more candidates survive the batch and \emph{vice versa}. 

\begin{algorithm}
  \SetKwInOut{Parameter}{Parameters}
  \KwData{cases := list of training cases in random order, candidates := population of classifiers}
  \Parameter{batch size, fitness threshold}
 \KwResult{Return a individual to be used as a parent }
 batches := create batches from the cases of a given batch size\;
 \While{True}{
   batch-sample := first batch from batches\;
   candidates := candidates with the fitness (number of corrects/ number of matches) > fitness threshold in the batch-sample\;
  \If{only one candidate exists in candidates}{
  	parent = candidate\;
	return parent;
   }
   delete batch-sample from batches; \\
   \If{batches is empty}{
	parent = a randomly selected candidate from candidates\;
	return parent;
    }
    
 }
 \caption{Batch-lexicase Selection}
\end{algorithm}

\section{Experiments \& Results}
We compare the effect of different parent selection methods described above on the rule discovery module of LCS. We focus on the generalization capability of the classifier especially in situations where there is only partial data available. Four binary classification problems described in Section 4.2 are used to test our hypothesis. To ensure that our results are statistically significant, we employed leave-group-out cross validation~\cite{MC}. A randomized portion of the data is held out in each trail as the test set and the classifier is trained on rest of the data. Accuracy over training and test set averaged over 50 independent runs are reported. 

Further we investigate the generalization capabilities of the rules generated by employing different parent selection methods when a sizable chunk of data (about 50\%) is unavailable or missing which is likely to happen in an online setting. To test this hypothesis we randomly discard a portion of data and analyze the effect on performance. We proceed to show that lexicase variants develop more generic rules that help in better generalization upon seeing fewer data.

\subsection{Hyperparameters}
Most of the hyperparameters for the UCS architecture are fixed at their standard values. The details of the fixed values are given in Table 1. A more detailed description and notation regarding the hyperparameters is given in ~\cite{XCS2}. GA subsumption is disabled. $N$, the maximum population size of the rules is set at $1000$. 

Generally in GA, fixed tournament sizes are used. Unlike in traditional GAs where the parents are chosen from the entire population, in LCS parents are chosen only from the correct set which is variable in size. In the case of LCS, a variable tournament size dependent on the current correct set size $|C|$ is implemented. A value of $0.4*|C|$ is proven to be robust\cite{Robust}. 

For batch-lexicase, batch size is set to a constant value of $100$ and the threshold for batch survival is kept constant at $0.9$. The sensitivity of batch-lexicase to parameter tuning is discussed further in section 4.4.
\begin{table}
  \caption{Hyperparameter settings fixed across all experiments}
  \label{tab:freq}
  \begin{tabular}{ccl}
    \toprule
    Module & Parameters\\
    \midrule
    Fitness & $\alpha = 0.1$, $\beta = 0.2$, $\nu = 5$\\
    GA & $\theta_{GA} = 25$, $\chi = 0.8$, $\mu = 0.04$\\
    Deletion & $\theta_{del} = 20$, $\delta = 0.1$\\
  \bottomrule
\end{tabular}
\end{table}

\subsection{The Multiplexer Problem}
The multiplexer problem is a boolean classification problem with $k + 2^{k}$ features. The first $k$ features/bits are called address bits which point to the location of bit from which the class of the data is to be read. An N-bit multiplexer has a total of $2^N$ data cases. Multiplexer is widely studied in the context of LCS and it was shown that LCS outperforms many machine learning methods in solving the multiplexer problem.~\cite{Mux}

\begin{figure*}
\centering     
\subfigure[Training Accuracy]{\label{fig:a}\includegraphics[width=58mm]{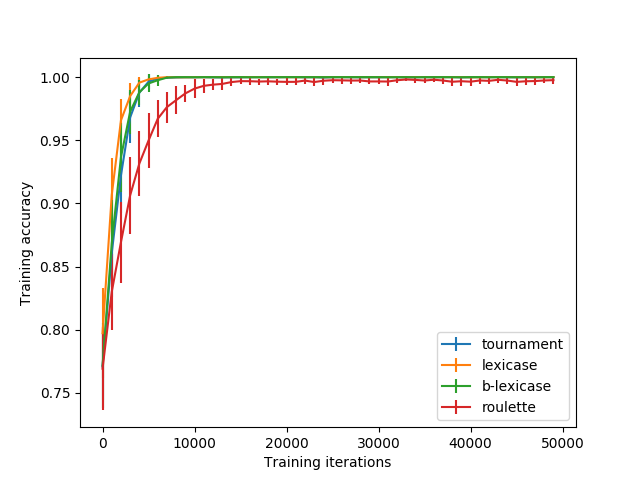}}
\subfigure[Test Accuracy]{\label{fig:b}\includegraphics[width=58mm]{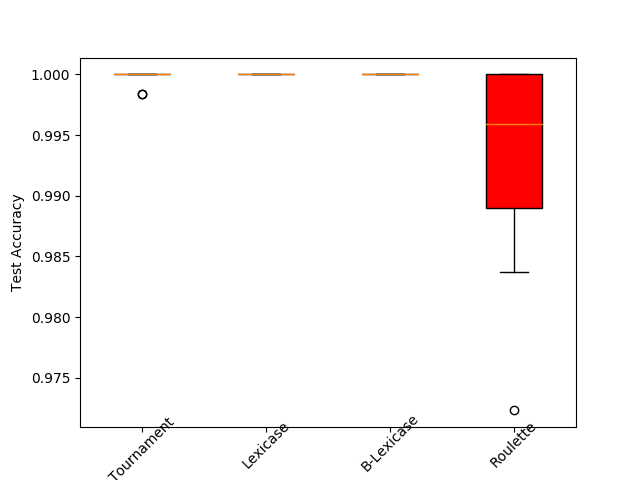}}
\subfigure[Number of rules generated]{\label{fig:c}\includegraphics[width=58mm]{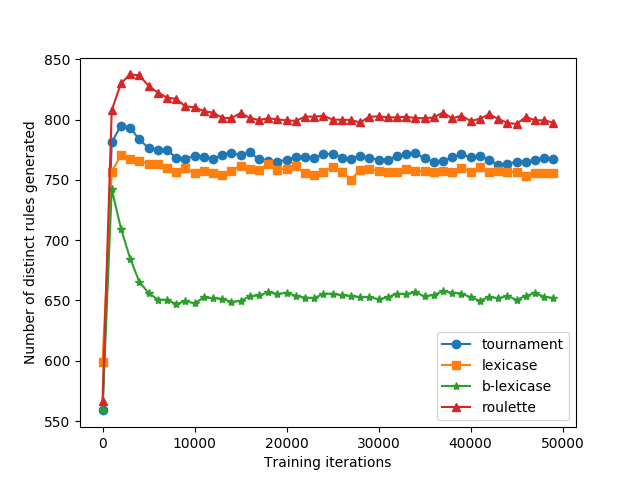}}
\caption{Results for a 11-bit multiplexer averaged over 50 independent runs. Standard deviation bars report fluctuations over 50 runs}
\end{figure*}

\begin{figure*}
\centering     
\subfigure[Training Accuracy]{\label{fig:d}\includegraphics[width=58mm]{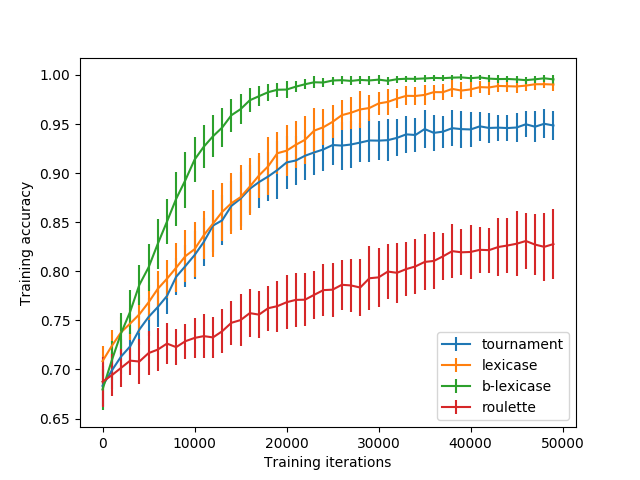}}
\subfigure[Test Accuracy]{\label{fig:e}\includegraphics[width=58mm]{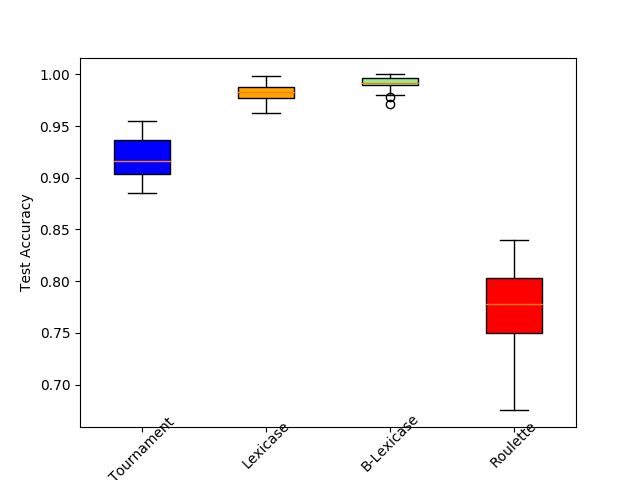}}
\subfigure[Number of rules generated]{\label{fig:f}\includegraphics[width=58mm]{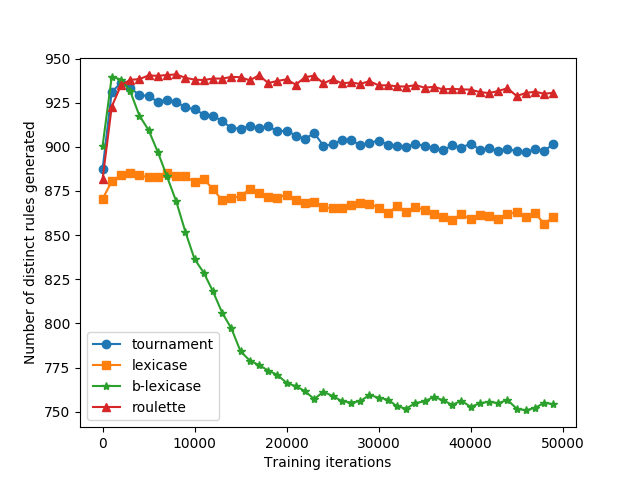}}
\caption{Results for a 20-bit multiplexer partial data averaged over 50 independent runs. Batch-lexicase has the highest accuracy on the test set with the least amount of fluctuation over 50 runs}
\end{figure*}

Figure 1(a) shows the training accuracy and Figure 1(b) shows the test accuracy on a 11-bit multiplexer for different selection methods. The train-test split used here is 70-30. All the four selection methods converge to an accuracy close to 1 on the training set and save for roulette wheel all the selection methods achieve perfect accuracy on the test set. Figure 1(c) plots the number of rules generated at various stages of training. Batch-lexicase selection generates fewer rules and gives identical performance compared to tournament selection. This could mean that the rule population generated by the batch-lexicase selection is more generic compared to the rest of the methods. Figure 3 shows the histogram of the rule distribution for a sample run of the algorithm where the bins correspond to the number of data instances correctly represented by a rule. From the figure we see that in tournament selection a majority of the rules describe fewer than 10 data instances while the batch-lexicase selection produces a rule distribution with each rule describing more data instances on average.

If the rules generated by the lexicase selection are indeed more generic then it could have a higher generalization ability. This hypothesis is further tested by considering a 20 bit multiplexer with partial data. Only 2000 data samples are considered out of $2^{20}$ possible data samples. Figure 2(a) and 2(b) show the plots for training and test accuracy using partial data for 20-bit multiplexer. Batch-lexicase and lexicase reach training accuracy of close to 1 even in a situation of sparse data. Batch-lexicase achieves a test accuracy of close to 1 outperforming all the other methods while generating fewer rules as shown in Figure 2(c).

\begin{figure}
\includegraphics[width=80mm]{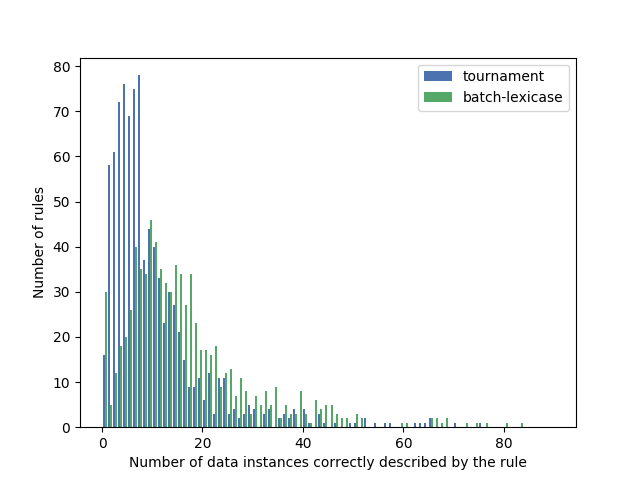}
\caption{Histogram of the rule distribution for the 11 bit multiplexer problem where the bins correspond to the number of data instances correctly represented by a rule. Batch-lexicase selection produces a rule distribution with each rule describing more data instances on average}
\end{figure}

\subsection{Other problems}
We repeated the analysis performed on the multiplexer on three other classification problems. In this study, we restrict ourselves to binary features. The parity problem which is another widely studied problem in LCS literature is studied. The other two problems are chosen from the UCI Machine Learning Repository~\cite{UCI}. The LED problem is chosen for its noisy nature and the car-evaluation problem for its imbalanced structure. The descriptions for each of the problems is given below. 

\subsubsection{The parity problem}
Parity problem is a boolean classification defined as follows: If the total number of ones in the features is odd the class is one and zero otherwise. In this case, we deal with a 10 bit parity problem.
\subsubsection{The LED problem}
The LED problem has ten categories each representing a digit from 0-9 and 7 binary features. This is a noisy dataset where each feature has a complementary value with a probability of 0.1. 

\subsubsection{The car-evaluation problem}
This dataset evaluates the quality of cars based on 21 binary features specifying various attributes into 4 classes ranging from unacceptable to good. This is a heavily unbalanced dataset.

 The training and test accuracy results for each of the problems are first reported using the entire dataset with a train-test split of 70-30. Then about $40\%$ of the data is discarded at random and the classifier is trained on $40\%$ of the data and the performance is evaluated using the rest of the data. The portion of data discarded is randomized at every trial. The results averaged over 50 independent runs are reported. Standard deviation bars are shown for each of  the plots reporting fluctuations across the 50 independent runs.

\begin{figure*}
\centering     
\subfigure[Training Accuracy on full data]{\label{fig:a}\includegraphics[width=58mm]{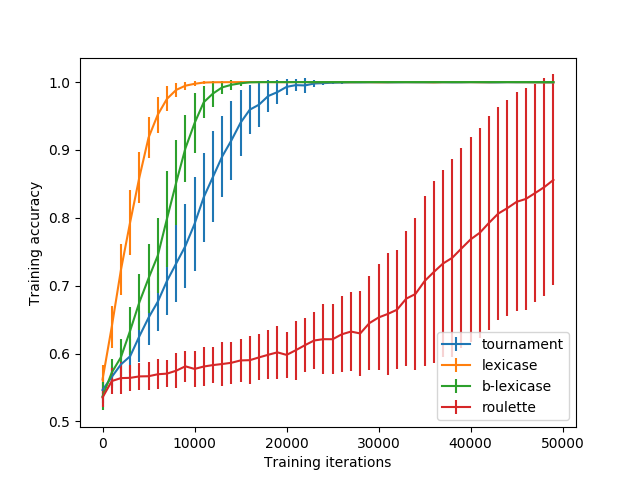}}
\subfigure[Test Accuracy on full data]{\label{fig:b}\includegraphics[width=58mm]{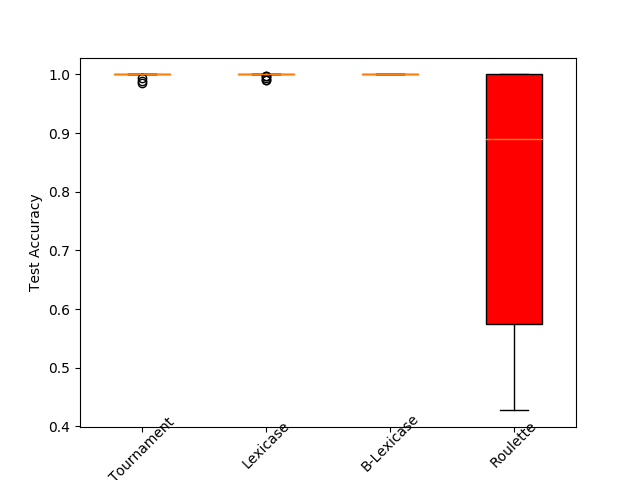}}
\subfigure[Number of rules generated]{\label{fig:c}\includegraphics[width=58mm]{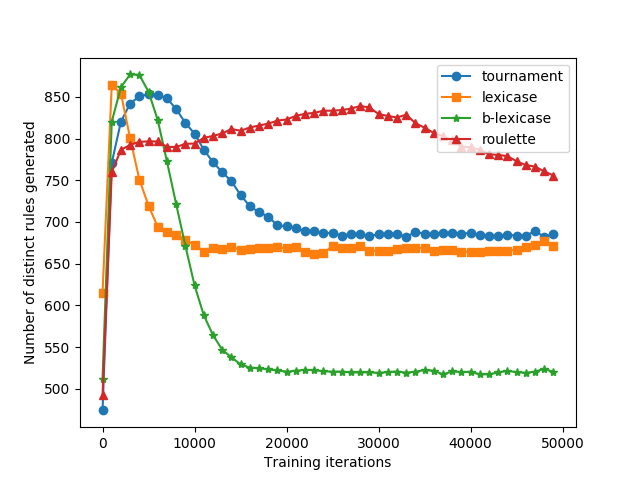}}
\subfigure[Training Accuracy on partial data]{\label{fig:d}\includegraphics[width=58mm]{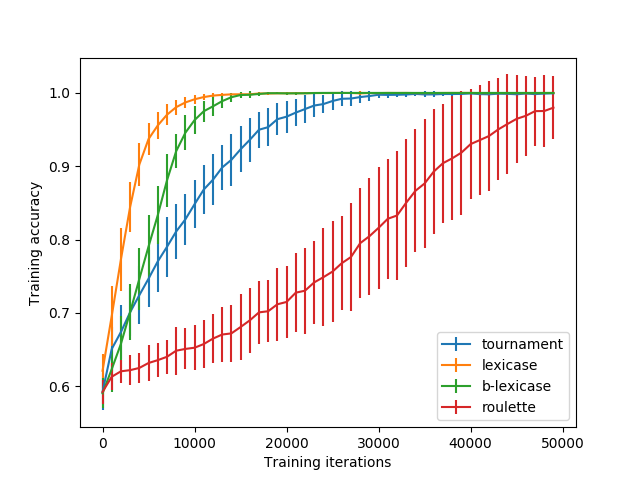}}
\subfigure[Test Accuracy on partial data]{\label{fig:e}\includegraphics[width=58mm]{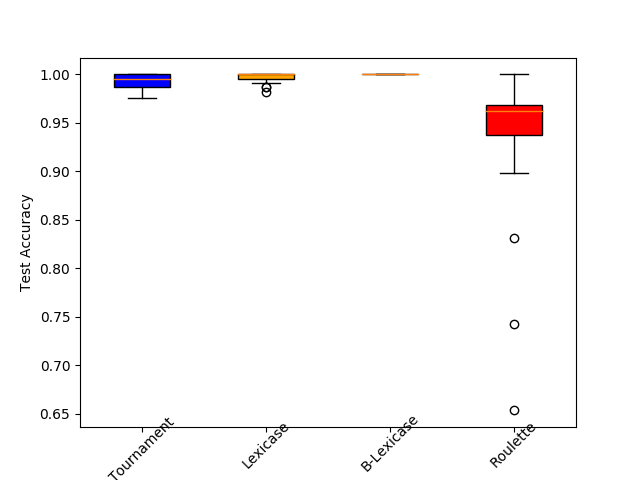}}
\subfigure[Number of rules generated]{\label{fig:f}\includegraphics[width=58mm]{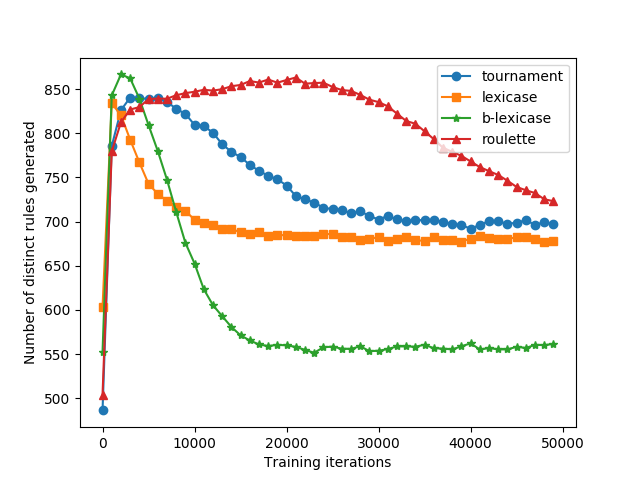}}
\caption{Results for the parity problem averaged over 50 independent runs. Batch-lexicase selection converges to perfect test accuracy even in situation of partial data}
\end{figure*}

We summarize the performance of various parent selection methods by providing the plots of training accuracy and test accuracy for each problem. Figure 4 shows the corresponding plots for the parity problem. As in the case of 11-bit multiplexer, most of the selection methods perform well when the entire data is made available. But in case of missing data, batch-lexicase selection converges to a higher training accuracy in fewer generations compared to the rest of the methods. The test accuracy is also highest for the batch-lexicase selection validating its generalization performance. Furthermore the standard deviation of the test accuracy is lower for batch-lexicase selection indicating lower amount of fluctuation and a more assured convergence to an optimal accuracy.

\begin{figure*}
\centering     
\subfigure[Training Accuracy on full data]{\label{fig:a}\includegraphics[width=58mm]{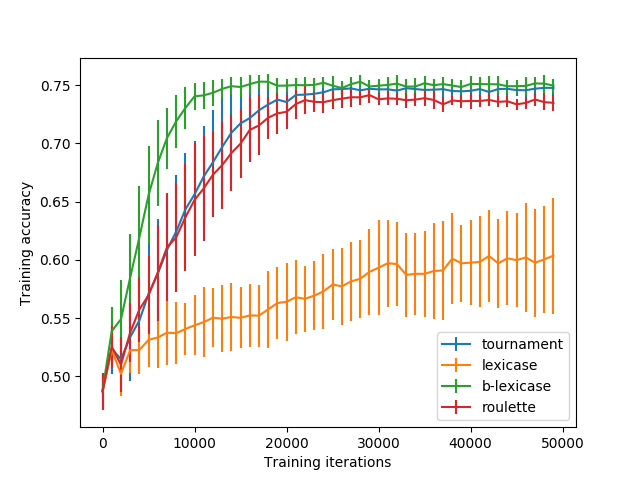}}
\subfigure[Test Accuracy on full data]{\label{fig:b}\includegraphics[width=58mm]{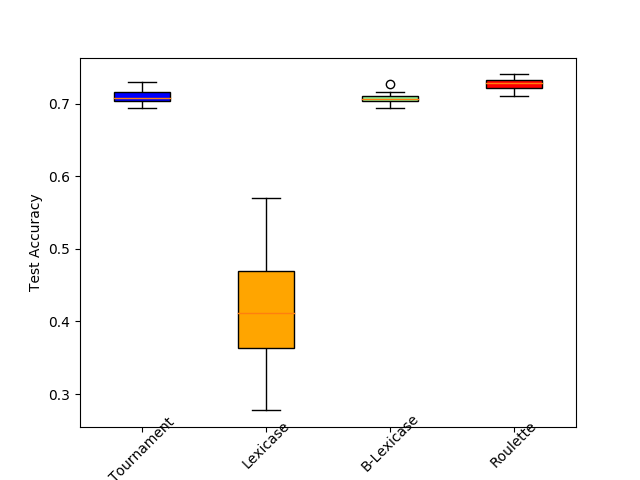}}
\subfigure[Number of rules generated]{\label{fig:c}\includegraphics[width=58mm]{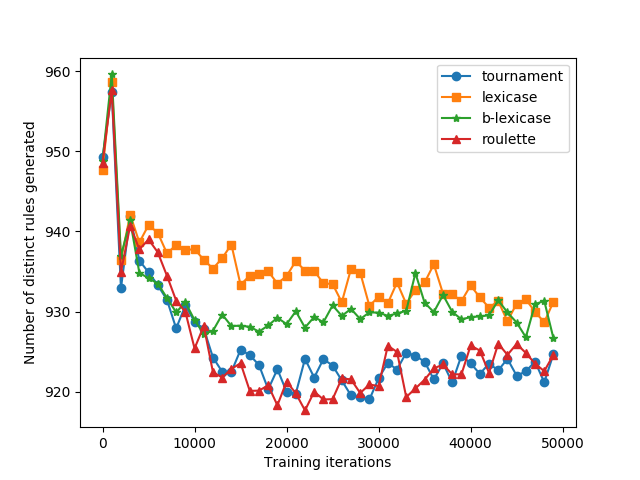}}
\subfigure[Training Accuracy on partial data]{\label{fig:d}\includegraphics[width=58mm]{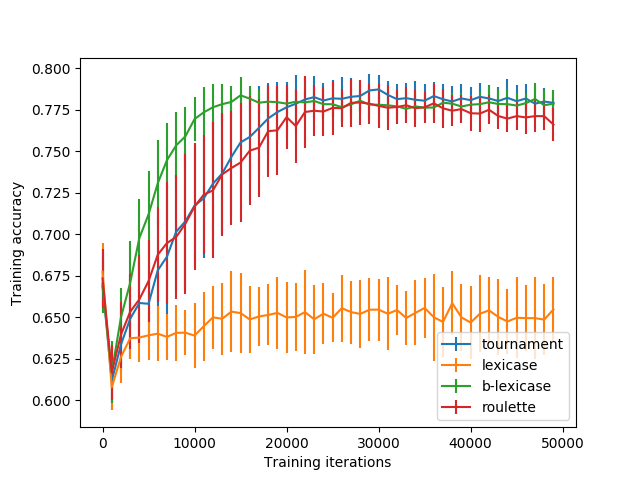}}
\subfigure[Test Accuracy on partial data]{\label{fig:e}\includegraphics[width=58mm]{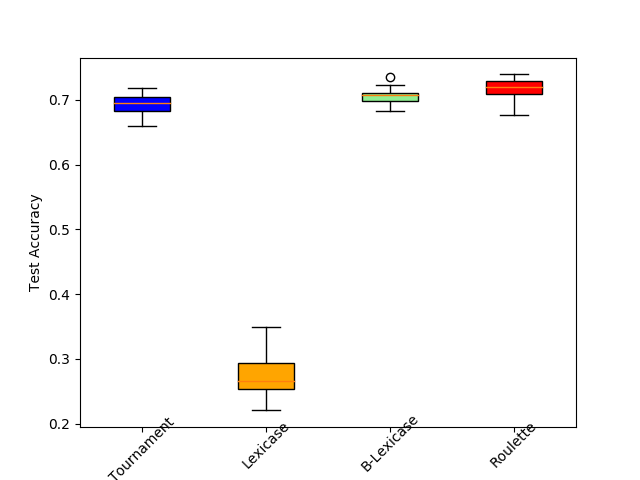}}
\subfigure[Number of rules generated]{\label{fig:f}\includegraphics[width=58mm]{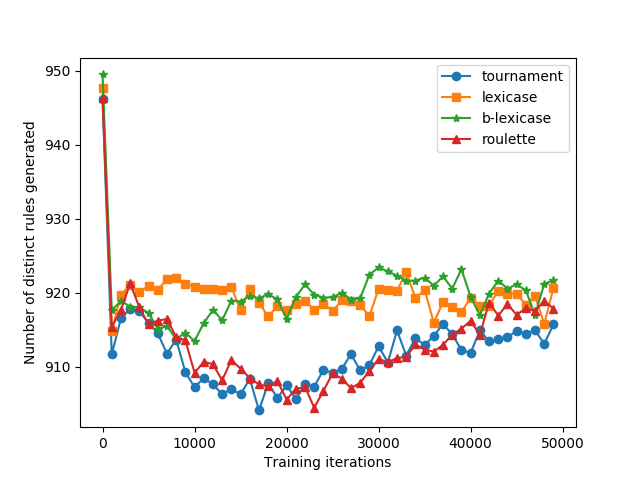}}
\caption{Results for the LED problem averaged over 50 independent runs. Batch-lexicase gives better generalization in the partial data situation and lexicase selection gives poor results}
\end{figure*}

Figure 5 shows the results for the LED problem. Interestingly lexicase selection gives the worst performance whereas batch-lexicase selection performs reasonably well. Roulette wheel selection performs the best of all the methods. The results in the case of missing data are similar to those of other problems. Batch-lexicase's performance is only second to roulette wheel selection. 

The LED problem clearly depicts the motivation behind the formulation of the batch-lexicase selection. Figure 6 shows a histogram of the rule distribution  based on the number of data instances correctly described by each rule. The plots are generated using the tournament selection but the plots generated using other selection methods are of a similar nature.  It can be seen from the figure that most of the rules describe fewer than 5 data instances. The results from the plot can be contrasted from an equivalent plot generated for the multiplexer problem in Figure 3.  As described in section 3.4, in case of situations where the rule representation is sparse, i.e, the number of data instances that can be represented by a single rule is small, lexicase selection is expected to give a poor performance which can be clearly seen in the results of the LED problem as shown in Figure 5. 

\begin{figure}
\includegraphics[width=82mm]{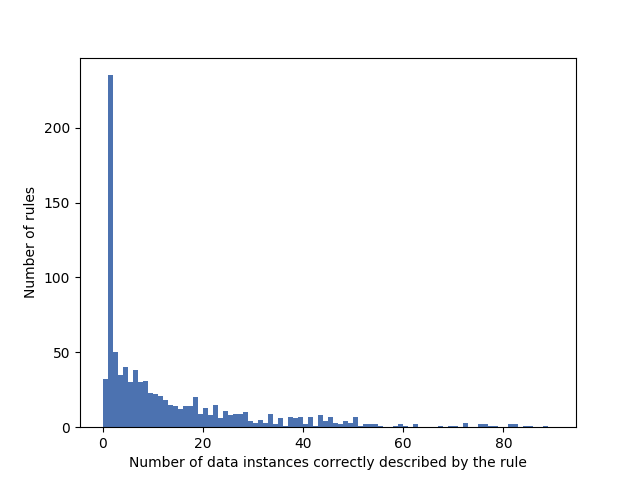}
\caption{Histogram of the rule distribution based on the number of data instances correctly described by each rule for the LED problem. The rule representation being sparse (most of the rules represent fewer than 5 data instances), the problem is not suitable for lexicase selection, providing a motivation for the formulation of batch-lexicase selection }
\end{figure}

\begin{figure*}
\subfigure[Training Accuracy on full data]{\label{fig:a}\includegraphics[width=58mm]{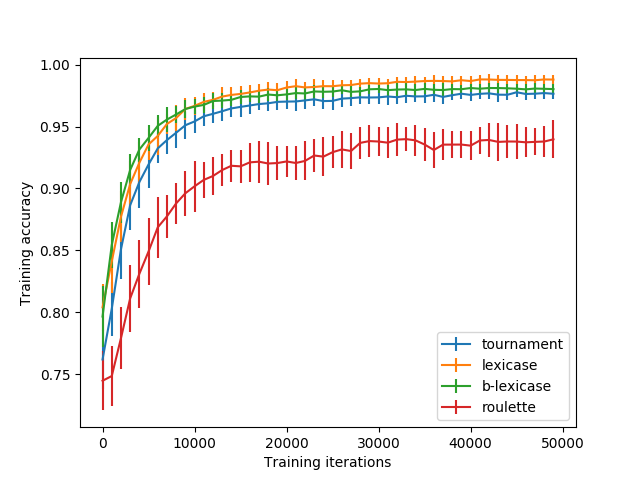}}
\subfigure[Test Accuracy on full data]{\label{fig:b}\includegraphics[width=58mm]{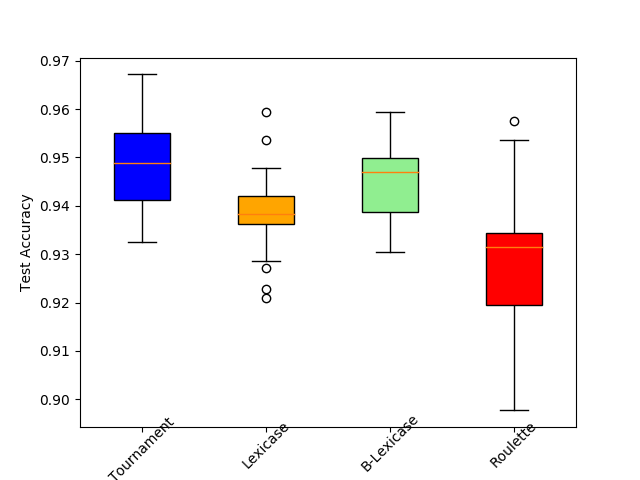}}
\subfigure[Number of rules generated]{\label{fig:c}\includegraphics[width=58mm]{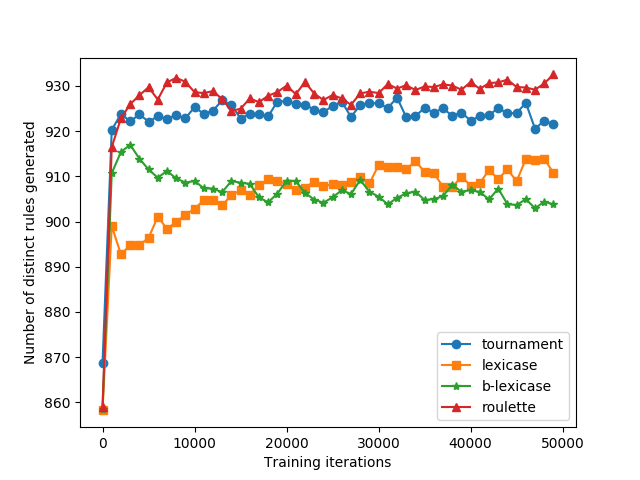}}
\subfigure[Training Accuracy on partial data]{\label{fig:d}\includegraphics[width=58mm]{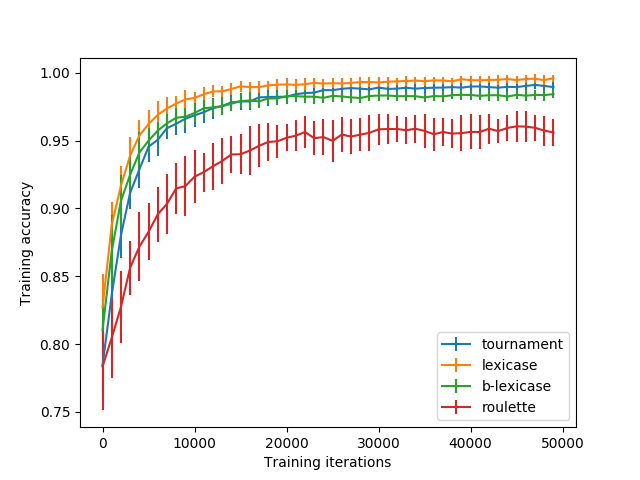}}
\subfigure[Test Accuracy on partial data]{\label{fig:e}\includegraphics[width=58mm]{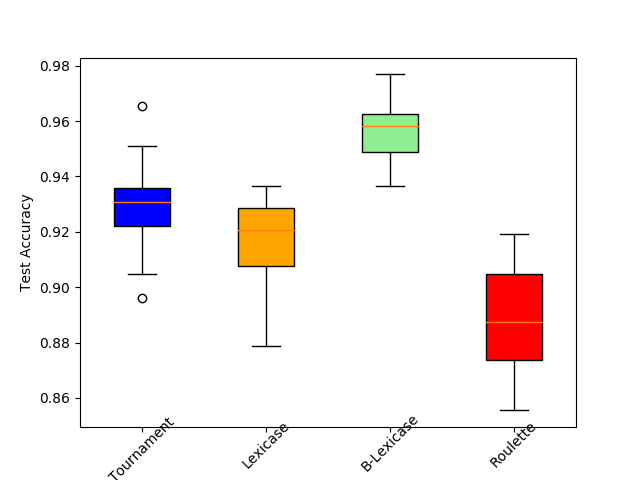}}
\subfigure[Number of rules generated]{\label{fig:f}\includegraphics[width=58mm]{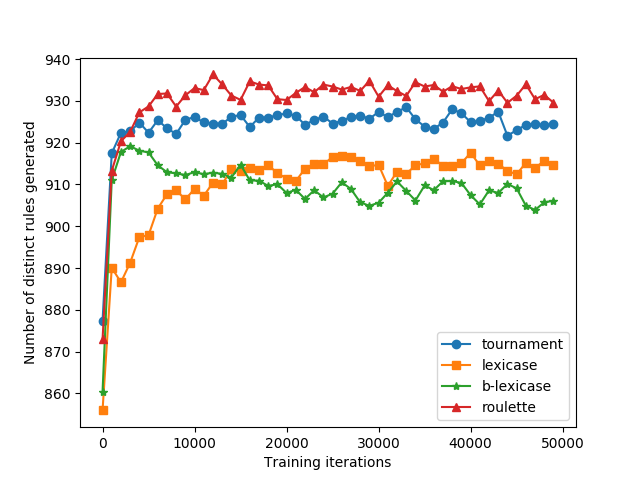}}
\caption{Results for the Car Evaluation problem averaged over 50 independent runs. Batch-lexicase gives better generalization on the test set in case of partial data}
\end{figure*}
Figure 7 shows the results obtained for the car-evaluation dataset. It should be noted that when the entire data is made available, tournament selection gives a slightly higher mean test accuracy compared to the batch-lexicase selection although there is high variance for tournament across different runs. When there is only part of the data available, batch-lexicase selection gives better generalization over the test set.

\subsection{Parameter Tuning}
In this experiment, the sensitivity of the batch-lexicase selection to its parameter tuning is investigated. The selection pressure is tuned by changing the batch size and the batch survival threshold. The learning curves are reported on the 20-bit multiplexer problem with partial data. Figure 8(a) shows the learning curves obtained on the training data by using batch size of 10, 50, 100, 200, 500 while keeping the batch survival threshold at 0.9. For the data size of 2000 samples, using a batch size of about 100, we converge to an optimal training accuracy. Figure 8(b) shows the learning curves obtained by using a batch survival threshold of 0.1, 0.3, 0.5, 0.7, 0.9 while keeping the batch size constant at 100. The batch survival threshold does not seem to be affecting the training accuracy by a huge margin in this problem. Although using a threshold of 0.9 gives the best training accuracy, it should be noted that the best hyperparameters of the batch-lexicase can be problem dependent just like the tournament size parameter for tournament selection. A batch size of $0.05 |N|$ to $0.1 |N|$, where $|N|$ is the size of the data and batch survival threshold of 0.9 is found to be robust for most problems.

\begin{figure*}
\centering     
\subfigure[Tuning of batch size]{\label{fig:a}\includegraphics[width=60mm]{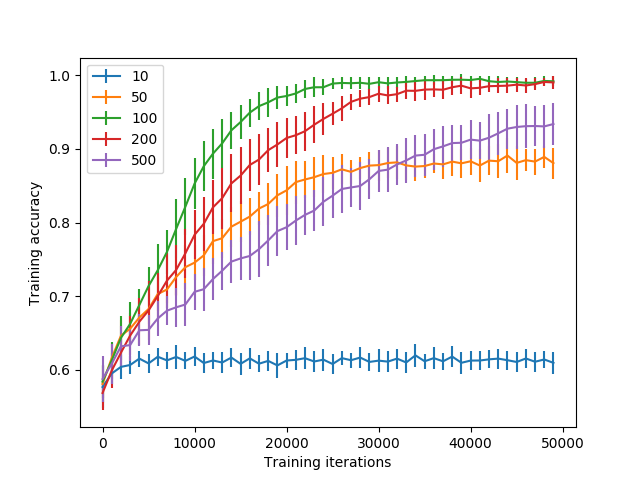}}
\subfigure[Tuning of batch survival threshold]{\label{fig:b}\includegraphics[width=60mm]{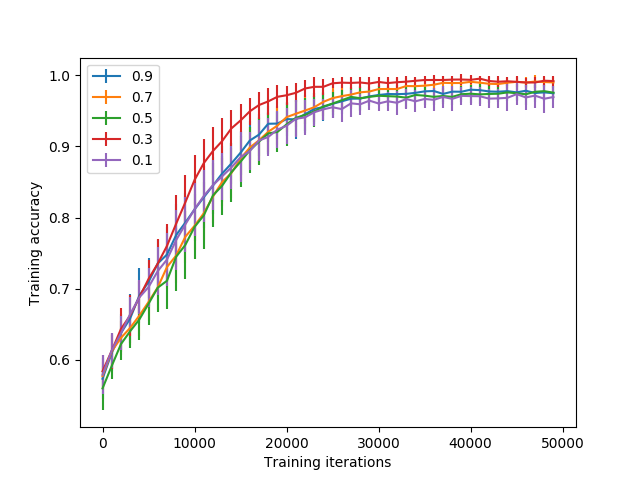}}
\caption{Parameter tuning results for the 20-bit multiplexer partial data averaged over 50 independent runs. Setting batch size too high or too low results in poorer performance}
\end{figure*}

\section{Conclusions}
In this paper we show that using batch-lexicase selection, a variant of lexicase selection, in LCS results in the creation of a more generalized rule representation of the data. This is particularly favorable in an online setting where the data available till an instant of time might not be representative of the true data distribution. Further we also showed that this generalization capacity holds in situations of missing data which could be of potential significance.

One of the shortcomings of incorporating the lexicase selection and its variants in the LCS is the increase in training time. Both lexicase and batch-lexicase selection results in the increase of training time by a factor of two in the experiments described above. However it is possible that faster training times can be achieved by exploring better data caching mechanisms in the algorithm.

Lexicase selection is found to be incompatible with certain data configurations as seen in the LED problem. The limitations of the batch-lexicase selection in dealing with different data configurations have not been found. A more rigorous theoretical analysis on the suitability of batch-lexicase in dealing with different kinds of problems needs to be done. A few studies\cite{Theory1},~\cite{Theory2} have performed a theoretical analysis on lexicase selection. The same could be extended for the batch-lexicase selection in the context of LCS and other forms of evolutionary machine learning. A deeper analysis on lexicase selection and its variants could result in improved parent selection methods with a fine-grained control over selection pressure. 

\section{ACKNOWLEDGMENTS}
This material is based upon work supported by the National Science Foundation under Grant No. 1617087. Any opinions, findings, and conclusions or recommendations expressed in this publication are those of the authors and do not necessarily reflect the views of the National Science Foundation.

\bibliographystyle{ACM-Reference-Format}
\bibliography{paper} 

\end{document}